\documentclass[letterpaper]{article} 
\usepackage{aaai2026}  
\usepackage{times}  
\usepackage{helvet}  
\usepackage{courier}  
\usepackage[hyphens]{url}  
\usepackage{graphicx} 
\usepackage{natbib}  
\usepackage{caption} 
\usepackage{algorithm}
\usepackage{algorithmic}

\usepackage{subfig}
\usepackage{amsmath}
\usepackage{amssymb}
\usepackage{booktabs}
\usepackage{multirow}
\usepackage[table]{xcolor}

\usepackage{newfloat}
\usepackage{listings}
\DeclareCaptionStyle{ruled}{labelfont=normalfont,labelsep=colon,strut=off} 
\floatstyle{ruled}
\newfloat{listing}{tb}{lst}{}
\floatname{listing}{Listing}
\nocopyright
\title{Lethe: Layer- and Time-Adaptive KV Cache Pruning for Reasoning-Intensive LLM Serving}
\author {
    Hui Zeng\textsuperscript{\rm 1, \rm 2},
    Daming Zhao\textsuperscript{\rm 3},
    Pengfei Yang\textsuperscript{\rm 1}\thanks{Corresponding author.},
    WenXuan Hou\textsuperscript{\rm 1},
    Tianyang Zheng\textsuperscript{\rm 1},
    Hui Li\textsuperscript{\rm 1},
    Weiye Ji\textsuperscript{\rm 1},
    Jidong Zhai\textsuperscript{\rm 3}
}
\affiliations {
    \textsuperscript{\rm 1}Xidian University, Xi'an, China\\
    \textsuperscript{\rm 2}Zhongguancun Academy, Beijing, China\\
    \textsuperscript{\rm 3}Tsinghua University, Beijing, China\\
    
    hzeng@stu.xidian.edu.cn, 
    cszdm@mail.tsinghua.edu.cn, 
    pfyang@xidian.edu.cn
    
}

\begin{document}

\maketitle

\begin{abstract}
Generative reasoning with large language models (LLMs) often involves long decoding sequences, leading to substantial memory and latency overheads from accumulating key-value (KV) caches. While existing KV compression methods primarily focus on reducing prefill memory from long input sequences, they fall short in addressing the dynamic and layer-sensitive nature of long-form generation, which is central to reasoning tasks. We propose \textbf{Lethe}, a dynamic KV cache management framework that introduces adaptivity along both the spatial and temporal dimensions of decoding. Along the spatial dimension, Lethe performs \emph{layerwise sparsity-aware allocation}, assigning token pruning budgets to each transformer layer based on estimated attention redundancy. Along the temporal dimension, Lethe conducts \emph{multi-round token pruning} during generation, driven by a \emph{Recency-Aware Selective Retention} (RASR) mechanism. RASR extends traditional recency-based heuristics by also considering token relevance derived from evolving attention patterns, enabling informed decisions about which tokens to retain or evict. Empirical results demonstrate that Lethe achieves a favorable balance between efficiency and generation quality across diverse models and tasks, increases throughput by up to 2.56×.
\end{abstract}

\begin{links}
    \link{Extended version}{https://arxiv.org/abs/2511.06029}
\end{links}

\begin{figure*}[htbp]
\centering
\includegraphics[width=7in]{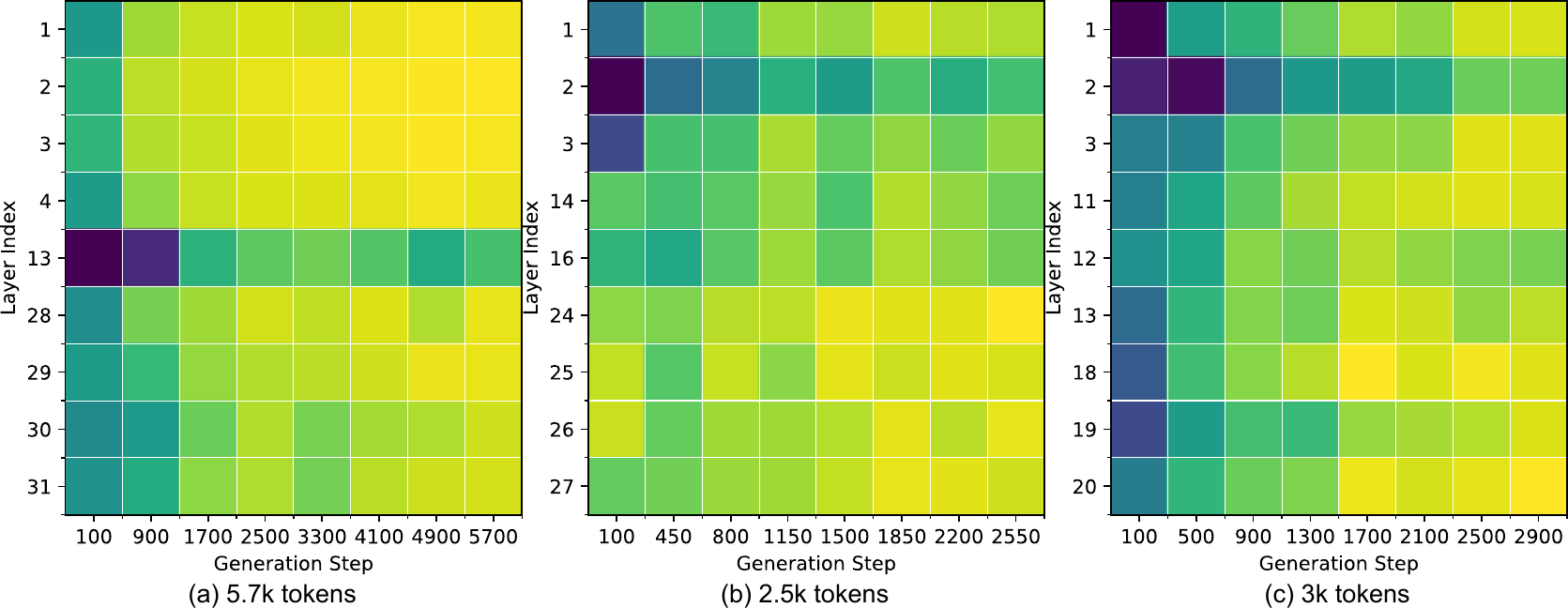}
\caption{Layerwise attention sparsity heatmaps over decoding steps for prompts from Math500 and MMLU using DeepSeek-R1-Distill-LLaMA-8B and Qwen-7B. Warmer colors indicate higher sparsity. Variations across time and layers motivate spatial-temporal adaptive cache management.}
\label{fig:three_images}
\end{figure*}

\section{Introduction}
Large Language Models have demonstrated impressive capabilities across a wide range of natural language understanding and generation tasks\cite{zhao2023survey}. Numerous studies have been proposed to optimize both the training and inference processes of LLMs, aiming to improve their efficiency and scalability\cite{10812976}. During inference, these models rely heavily on the \textit{Key-Value (KV) Cache} mechanism to store intermediate representations—specifically the key and value tensors from each transformer layer—in order to enable efficient autoregressive generation\cite{vaswani2017attention}. However, in many generation-intensive scenarios, especially those involving \textit{Chain-of-Thought (CoT)}\cite{wei2022chain,zhang2022automatic} prompting, the memory footprint of the KV cache grows rapidly with each decoding step. Unlike fixed-length prompts used in instruction tuning or retrieval-augmented generation, CoT-style generation introduces a unique challenge: tokens are not statically provided but are dynamically generated across multiple intermediate reasoning steps, leading to unpredictable and often lengthy sequences during inference.

This dynamic growth exacerbates the memory bottleneck in long-context scenarios, making existing optimizations—largely designed for compressing static prompts—insufficient. In particular, strategies that focus exclusively on reducing the prompt’s contribution to KV cache size fail to account for the cumulative and evolving nature of the cache during generation. Moreover, not all intermediate tokens contribute equally to final answers; overemphasis on historically high-attention tokens can mislead later predictions. This motivates the need for a principled mechanism to manage memory \textit{dynamically} during the decoding process.

There exist numerous methods\cite{zhang2023h2o,cai2024pyramidkv,hooper2024kvquant} and benchmarks\cite{zhang2024inftybench,bai2023longbench,an2023eval} aimed at addressing the challenges posed by long-context processing and evaluating the effectiveness of proposed solutions. While many approaches have been developed to reduce memory pressure from long contexts, they often rely on static assumptions that limit their applicability in dynamic generation scenarios. Specifically, these methods typically focus on optimizing the initial input alone and perform a one-time truncation or pruning after the prefill phase. Such strategies overlook two critical aspects in real-world chain-of-thought generation tasks:

First, there is significant \textbf{layerwise heterogeneity} in attention usage. While prior work such as PyramidKV~\cite{yang2024pyramidinfer} assumes monotonic patterns to allocate token budgets, we find that token importance varies non-monotonically across layers in reasoning tasks, making static allocation suboptimal. Second, \textbf{temporal inconsistency} in token relevance presents an equally important challenge. During autoregressive generation, attention distributions evolve with each decoding step, and the contextual importance of tokens fluctuates dynamically. Tokens that were once crucial may quickly become obsolete, while others gain importance as reasoning progresses. Static compression or one-time pruning fails to accommodate such shifts. 

Therefore, a temporal adaptation mechanism is necessary—one that continuously re-evaluates token utility as the generation unfolds. To address these challenges, we introduce \textbf{Lethe}, a memory-efficient inference framework that dynamically prunes the KV cache along both spatial and temporal dimensions. Lethe introduces two key techniques: (1) a layerwise sparsity estimator to allocate token budgets based on runtime attention patterns, enabling adaptive per-layer pruning; and (2) \textbf{Recency-Aware Selective Retention (RASR)}, a mechanism that incrementally prunes tokens based on both attention history and recent activity, supporting multi-round pruning during decoding.

The combination of these two strategies allows Lethe to significantly reduce memory overhead while preserving, and in some cases even improving, inference accuracy and throughput.

This paper makes the following key contributions:

\begin{itemize}
\item We reveal the layerwise and temporal variation in KV cache utility during transformer inference, showing that different layers contribute unevenly to output semantics and that relevance shifts over time. 
    
\item We propose \textbf{Lethe}, a memory-efficient inference framework that jointly addresses spatial and temporal inefficiencies. In the spatial dimension, Lethe employs a runtime \textbf{sparsity estimator} to adaptively allocate token budgets across transformer layers. In the temporal dimension, Lethe introduces a \textbf{Recency-Aware Selective Retention (RASR)} strategy, which performs iterative cache pruning based on a combination of attention history and recent usage, allowing context-sensitive memory reduction without compromising semantic fidelity.

\item We evaluate Lethe across multiple models and CoT-style tasks. Lethe achieves up to \textbf{2.56×} higher throughput and \textbf{91.7\%} reduction in KV cache memory usage compared to full caching, while maintaining or surpassing the accuracy of state-of-the-art baselines such as H2O, StreamingLLM, and PyramidKV.

\end{itemize}


\begin{figure*}[htbp]
\centering
\includegraphics[width=7in]{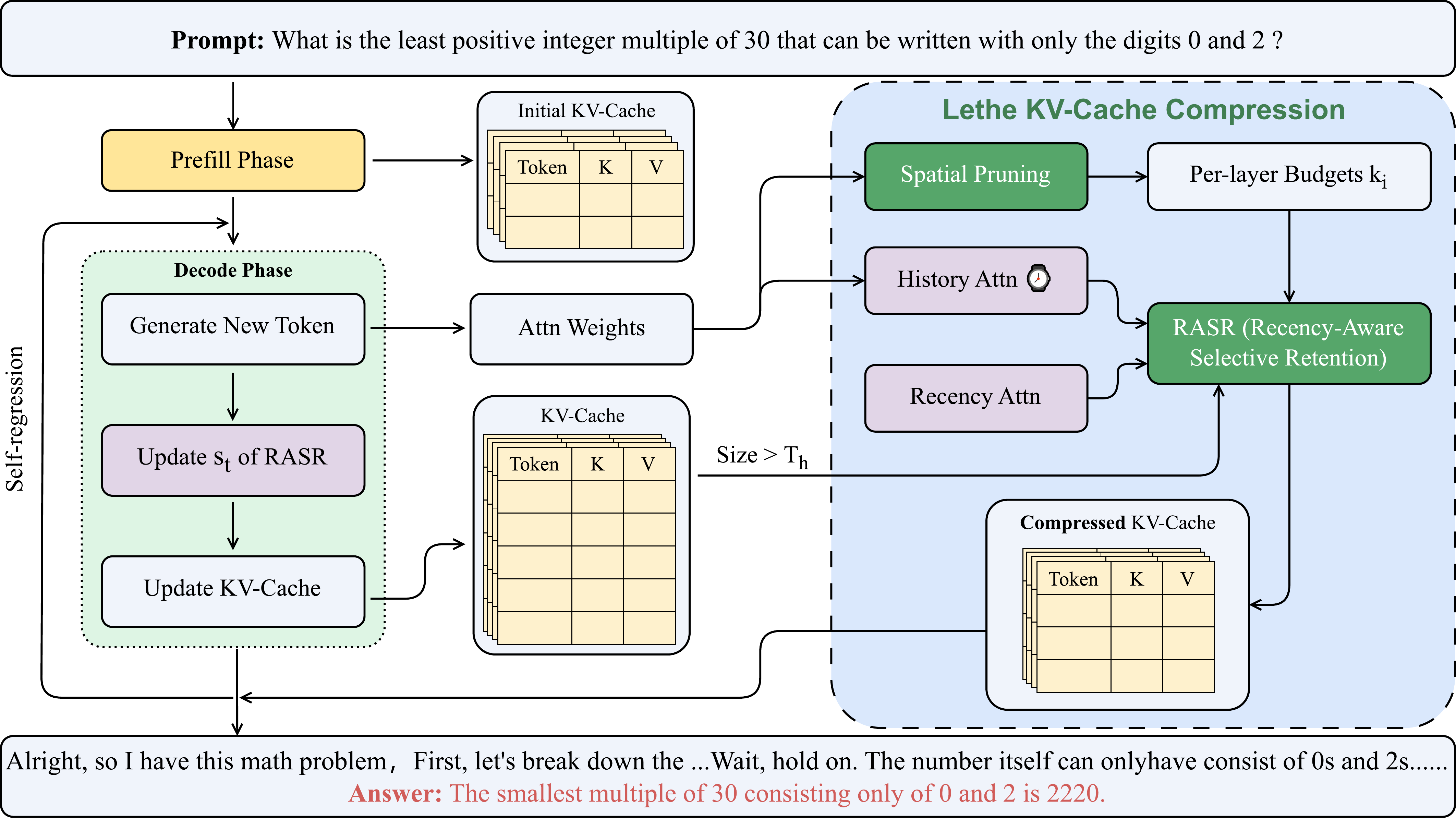}
\caption{System overview of Lethe. During decoding, Lethe prunes the growing KV cache using layerwise sparsity estimation and attention-based ranking, reducing memory and computation without sacrificing quality.}
\label{fig_overview}
\end{figure*}
\section{Related Work}
\label{relatedwork}

\subsection{Efficient Attention and System Optimizations}

To address the scalability limits of self-attention in autoregressive LLMs, prior work has explored both algorithmic and system-level optimizations.

On the algorithmic side, methods like Multi-Query Attention (MQA)~\cite{shazeer2019fast}, Grouped-Query Attention (GQA)~\cite{ainslie2023gqa}, and Multi-Head Latent Attention (MLA)~\cite{liu2024deepseek} reduce redundancy by sharing KV projections or introducing latent representations. Other approaches approximate attention using kernel methods~\cite{choromanski2020rethinking, peng2021random}, low-rank projections~\cite{wang2020linformer}, or block-sparsity~\cite{child2019generating, zaheer2020big}.

System-level solutions like FlashAttention~\cite{dao2022flashattention, dao2023flashattention, shah2024flashattention}, PagedAttention~\cite{kwon2023efficient}, and InfiniGen~\cite{298683} improve efficiency via fused kernels, memory paging, or KV prefetching. However, most of these methods retain all cached tokens, limiting scalability under long-context generation.

In contrast, Lethe dynamically prunes KV entries based on attention sparsity and temporal utility, enabling finer-grained memory control during decoding.

\subsection{Quantization of KV Cache}

Quantization offers a complementary way to reduce KV cache size by compressing stored representations~\cite{lin2024qserve,liu2024kivi,hooper2024kvquant,tao2024asymkv,duanmu2024skvq,yue2024wkvquant,tao2025cocktail}. 

KIVI~\cite{liu2024kivi} and KVQuant~\cite{hooper2024kvquant} apply adaptive granularity across tokens and channels. FlexGen~\cite{pmlr-v202-sheng23a} combines 4-bit quantization with offloading for constrained setups. AsymKV~\cite{tao2024asymkv} and SKVQ~\cite{duanmu2024skvq} exploit asymmetric precision and structured grouping, while CocktailQuant~\cite{tao2025cocktail} mixes bitwidths across layers or regions.

While these methods reduce storage per token, Lethe focuses on reducing the number of tokens themselves. The two are complementary: Lethe can be layered on top of quantized caches for compounded memory savings.

\subsection{Cache Eviction Strategies}

Evicting unimportant tokens has emerged as a key strategy for memory-efficient inference. One-shot methods such as H2O~\cite{zhang2023h2o}, SNAP-KV~\cite{li2024snapkv}, and Scissorhands~\cite{liu2023scissorhands} apply heuristics or saliency-based rules. Attention-guided methods like DMC~\cite{nawrot2024dynamic}, FastGen~\cite{ge2023model}, and Quest~\cite{tang2024quest} analyze attention scores or query activations.

Recent approaches consider structural trends: PyramidKV~\cite{cai2024pyramidkv} exploits layerwise attention shifts; Keyformer~\cite{adnan2024keyformer} keeps only high-impact keys; and StreamingLLM~\cite{xiao2023efficient} retains early attention sinks for long contexts.

Lethe differs by jointly modeling spatial and temporal dynamics. It allocates layer-specific budgets using sparsity estimates and applies recency-aware pruning during decoding. This unified, adaptive strategy better supports reasoning-intensive tasks where token utility evolves non-uniformly across layers and time.

\begin{figure*}[htbp]
\centering
\includegraphics[width=5.5in]{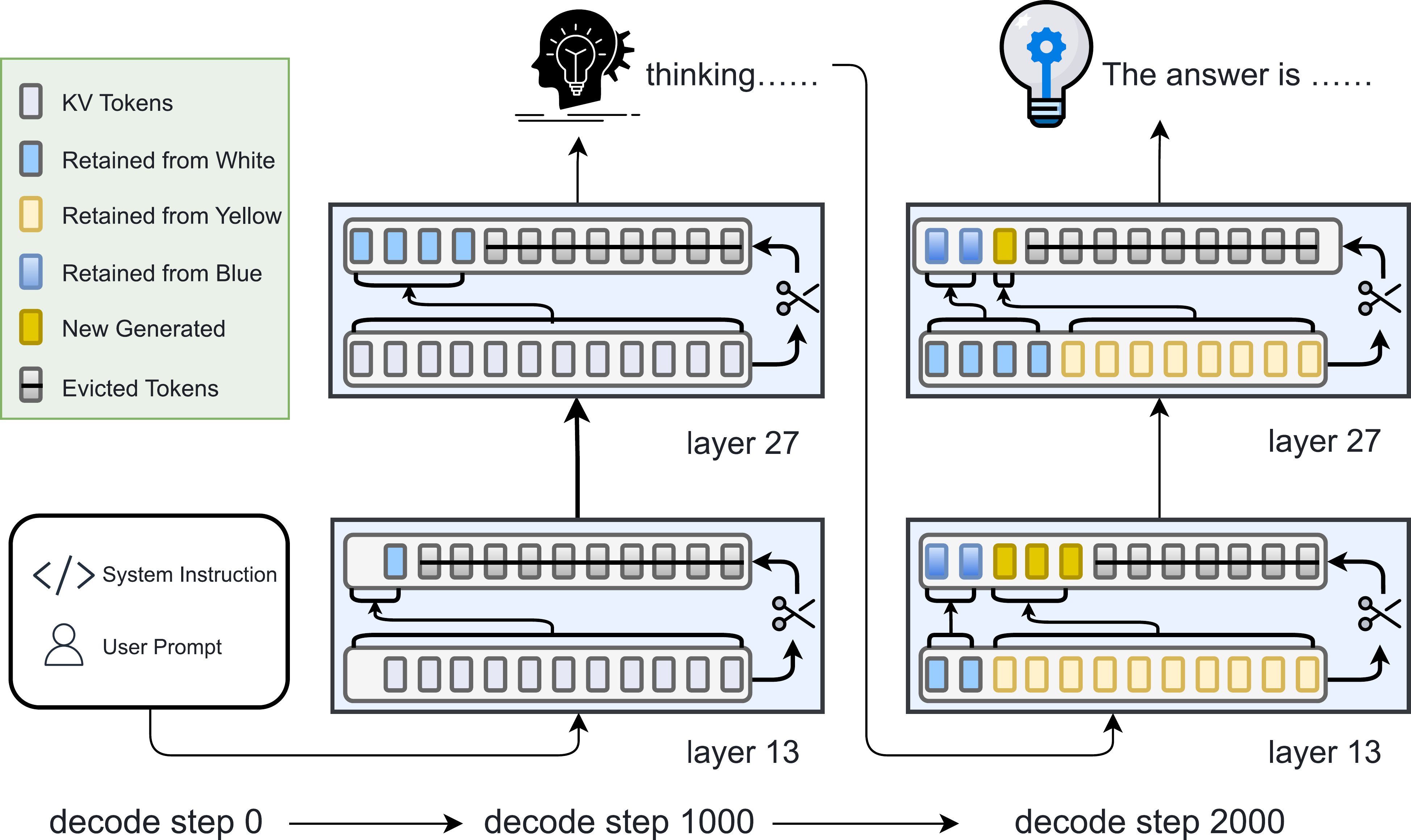}
\caption{Visualization of Lethe’s dual-dimensional KV cache pruning. Lethe prunes adaptively across layers and steps, reducing memory and computation in long-context decoding.}
\label{fig_lethe_illustration}
\end{figure*}

\section{Empirical Observations on Attention Sparsity}

\subsection{Attention Sparsity as a Basis for Fine-Grained Cache Management}

Prior work~\cite{zhang2023h2o} shows that attention weights correlate with token utility in generation, suggesting that not all tokens are equally important. This motivates selectively retaining high-importance tokens to reduce KV cache memory without sacrificing performance.

To formalize this, we quantify \emph{attention sparsity}—the degree to which attention is concentrated on a few tokens. A sparse distribution implies only a small subset of tokens are critical at each step.

We adopt the \textbf{Hoyer sparsity metric}~\cite{hoyer2004non}, a continuous, scale-invariant measure suitable for evaluating attention patterns across layers and models. Let $\mathbf{a} \in \mathbb{R}^n$ be a non-negative attention score vector. Its Hoyer sparsity is defined as:
\begin{equation}
\text{Sparsity}(\mathbf{a}) = \frac{\sqrt{n} - \frac{\|\mathbf{a}\|_1}{\|\mathbf{a}\|_2}}{\sqrt{n} - 1}
\end{equation}
This yields values in $[0,1]$, with higher values indicating more peaked (i.e., selective) attention. Attention sparsity thus provides a principled basis for fine-grained KV cache management.

\subsection{Layerwise Variability in Attention Sparsity}

Attention sparsity is not uniform across layers. Allocating KV cache equally across layers can be inefficient: sparse layers may be over-provisioned, while dense ones suffer from under-allocation.

While PyramidKV~\cite{cai2024pyramidkv,yang2024pyramidinfer} assumes a pyramidal sparsity trend—lower layers being dense and upper layers sparse—our analysis reveals this structure does not hold in reasoning models used with CoT prompts.

Figure~\ref{fig:three_images}(a--c) shows layerwise sparsity heatmaps for three Math500 prompts. For LLaMA-8B (a), both early and late layers are sparse, while mid-layers are dense—contradicting the pyramid assumption. Qwen-7B (b, c) shows varied trends: in (b), sparsity increases steadily; in (c), it fluctuates unpredictably. These results highlight prompt-specific and non-monotonic sparsity patterns.

Such variability indicates that fixed, layer-agnostic cache allocation strategies—uniform or pyramidal—are inherently suboptimal, particularly for complex reasoning tasks. An adaptive strategy is needed to match the diverse and dynamic sparsity across layers.

\subsection{Temporal Dynamics in Attention Patterns During Decoding}

Attention sparsity also evolves over decoding steps. Unlike standard text generation, CoT-style reasoning involves multi-step chains of thought, making the model's attention patterns temporally dynamic.

Figure~\ref{fig:three_images}(c) shows that early layers begin with low sparsity but become increasingly selective, while others fluctuate (e.g., layer 13 in (a)). This temporal evolution implies that fixed per-layer budgets fail to reflect shifting token importance.

In summary, both spatial (layerwise) and temporal (stepwise) variations in attention sparsity demand a dynamic, adaptive approach to KV cache management for efficient and high-quality inference.

\section{Lethe}
\label{Lethe}
\subsection{System Overview}
\label{Lethe-1}

Lethe is an inference-time memory management framework for large language models, designed to reduce memory and computation by dynamically pruning the key-value (KV) cache. As shown in Figure~\ref{fig_overview}, Lethe operates during autoregressive decoding: after an initial prefill stage, the KV cache grows with each generated token. To avoid memory overflow, Lethe monitors cache size and triggers pruning once a configurable threshold is exceeded. Guided by attention sparsity, Lethe selectively evicts low-utility tokens on a per-layer basis, preserving model quality while improving efficiency.

\subsection{Spatial Key-Value Cache Pruning via Layerwise Sparsity Estimation}
\label{Lethe-LST}

Lethe introduces spatially adaptive KV pruning, allocating memory differently across transformer layers based on measured sparsity. Unlike fixed-budget or preallocated schemes, Lethe adapts retention dynamically, improving utilization in deep models where attention varies significantly across layers.

Token importance is computed by aggregating attention weights across heads and batches. Let $\mathcal{A}^{(l)} \in \mathbb{R}^{B \times H_Q \times Q \times K}$ be the raw attention tensor; Lethe collapses it as:
\begin{equation}
\mathbf{s}^{(l)} = \sum_{b=1}^{B} \sum_{h=1}^{H_Q} \sum_{q=1}^{Q} \mathcal{A}^{(l)}_{b,h,q,:}
\end{equation}

This head-invariant scoring supports architectures like GQA/MQA by avoiding duplicated key expansion:
\begin{equation}
\tilde{\mathbf{K}} = \mathrm{repeat}(\mathbf{K}, \text{repeats}=H_Q / H_{KV}, \text{axis}=1)
\end{equation}

Tokens are sorted by $\mathbf{s}^{(l)}$ and divided into $D$ segments. Lethe identifies the first segment where attention drops sharply:
\begin{equation}
\frac{\mathbf{v}_{\text{top}}[0]}{\mathbf{v}_{\text{top}}[k^*]} \leq \tau
\end{equation}

The selected top-$k$ tokens, combined with recent and sink tokens, are retained. If no breakpoint is found, Lethe conservatively delays pruning. Algorithm~\ref{alg:lethe_shrink} formalizes this.

\begin{algorithm}[htbp]
\caption{\textsc{Segmented Attention-Based Token Shrinking (Lethe)}}
\label{alg:lethe_shrink}
\begin{algorithmic}[1]
\REQUIRE
    Attention score vector $s \in \mathbb{R}^{K}$ from layer $l$, \\
    Number of segments $D$, \\
    Threshold parameter $\tau > 1$, \\
    Sink length $s_\text{len}$, \\
    Recent window size $r$, \\
    Initial eviction threshold $L_{\text{evict}}$

\ENSURE
    Token retention indices $\mathcal{T} \subseteq \{0, 1, \ldots, K-1\}$, \\
    Updated eviction threshold $L_{\text{evict}}$

\STATE $\text{top\_values}, \text{top\_indices} \leftarrow \text{TopK}(s, K)$
\STATE $\text{cut\_points} \leftarrow \left\{ \left\lfloor \frac{K \cdot d}{D} \right\rfloor \mid d = 1, \dots, D-1 \right\}$
\STATE $\text{breakpoint} \leftarrow -1$

\FOR{each $c \in \text{cut\_points}$}
    \STATE $v_{\text{head}} \leftarrow \text{top\_values}[0]$
    \STATE $v_{\text{cut}} \leftarrow \text{top\_values}[c]$
    \IF{$\frac{v_{\text{head}}}{v_{\text{cut}}} \leq \tau$}
        \STATE $\text{breakpoint} \leftarrow c$
        \STATE \textbf{break}
    \ENDIF
\ENDFOR

\STATE $\text{sink\_indices} \leftarrow \{0, 1, \dots, s_\text{len}-1\}$
\STATE $\text{recent\_indices} \leftarrow \{K - r, \dots, K-1\}$

\IF{$\text{breakpoint} \geq 0$}
    \STATE $\text{salient\_indices} \leftarrow \text{top\_indices}[:\text{breakpoint}]$
    \STATE $L_{\text{evict}} \leftarrow \max(L_{\text{evict}}, \text{breakpoint} + r)$
\ELSE
    \STATE $L_{\text{evict}} \leftarrow 2 \cdot L_{\text{evict}}$
\ENDIF

\RETURN $\mathcal{T}, L_{\text{evict}}$
\end{algorithmic}
\end{algorithm}

\subsection{Temporal Memory Management with Recency-Aware Selective Retention (RASR)}
\label{Lethe-LFM}

To address the temporal growth of key-value (KV) caches in autoregressive decoding, Lethe introduces \textbf{Recency-Aware Selective Retention} (RASR), which prunes outdated or low-utility tokens based on attention history. Unlike conventional heuristics (e.g., LRU, LFU), RASR integrates both \emph{recency} and \emph{significance} derived from attention dynamics.

At each decoding step $t$, Lethe maintains a score vector $\mathbf{s}_t \in \mathbb{R}^{L_t}$ for all cached tokens:
\begin{equation}
\mathbf{s}_t = \gamma \cdot \mathbf{s}_{t-1} + \sum_{h=1}^{H} \sum_{i=1}^{q} \sum_{j=1}^{k} \mathbf{A}^{(t)}_h(i, j),
\end{equation}
where $\gamma \in (0, 1)$ controls decay, and $\mathbf{A}^{(t)}_h(i,j)$ denotes attention weight from head $h$ to token $j$.

Tokens are periodically ranked by a combination of $\mathbf{s}_t$ and their age. Those below a dynamic threshold are evicted, except for recent and boundary tokens. This allows Lethe to gradually forget stale context while preserving salient history.

RASR complements spatial pruning (Section~\ref{Lethe-LST}) by operating along the time axis. Together, they enable lightweight, adaptive, and repeated memory reduction with minimal overhead. Empirical results show RASR reduces memory and improves output quality in long-context inference.

{
\renewcommand{\arraystretch}{0.7}
\begin{table*}[ht]
\centering
\begin{tabular}{lccccccccc}
\toprule
\multirow{7}{*}{Method} & \multicolumn{8}{c}{MMLU} & \multirow{7}{*}{math500} \\
\cmidrule(lr){2-9} 
 & \rotatebox{45}{ab. algebra} & \rotatebox{45}{anatomy} & \rotatebox{45}{astronomy} & \rotatebox{45}{bus. ethics} & \rotatebox{45}{clin. knowl.} & \rotatebox{45}{coll. biology} & \rotatebox{45}{coll. chem.} & \rotatebox{45}{coll. cs} & \\
\midrule
\multicolumn{10}{c}{DeepSeek-R1-Distill-Qwen-7B} \\ 
\midrule
FullKV & 85.00 & 51.85 & 72.37 & 64.00 & 61.89 & 60.42 & 58.00 & 78.00 & 86.40 \\
H2o & 81.00 & 42.96 & 72.37 & 62.00 & \textbf{63.40} & 57.64 & 52.00 & 63.00 & 68.00 \\
StreamingLLM & 86.00 & 45.19 & \textbf{77.63} & 65.00 & 63.02 & 58.33 & 57.00 & 66.00 & 67.60 \\
PyramidKV & 84.00 & \textbf{51.11} & 75.00 & 62.00 & 60.38 & 63.89 & 49.00 & 64.00 & 58.40 \\
\rowcolor{gray!10}
\textbf{Lethe(ours)} & \textbf{87.00} & 48.89 & 73.68 & \textbf{66.00} & 63.02 & \textbf{65.28} & \textbf{61.00} & \textbf{74.00} & \textbf{85.40} \\
\midrule
\multicolumn{10}{c}{DeepSeek-R1-Distill-Qwen-32B} \\
\midrule
FullKV & 92.00 & 74.81 & 94.08 & 83.00 & 87.17 & 97.92 & 65.00 & 92.00 & 88.00 \\
H2o & 92.00 & 79.25 & 90.13 & 77.00 & \textbf{87.92} & \textbf{95.14} & 65.00 & 81.00 & 69.40 \\
StreamingLLM & 93.00 & 80.74 & 93.42 & 82.00 & 86.42 & 94.44 & \textbf{69.00} & 86.00 & 74.00 \\
PyramidKV & 79.00 & 80.00 & \textbf{94.74} & 81.00 & \textbf{87.92} & 92.36 & 51.00 & 75.00 & 72.60 \\
\rowcolor{gray!10}
\textbf{Lethe(ours)} & \textbf{94.00} & \textbf{81.48} & 92.11 & \textbf{84.00} & 87.17 & \textbf{95.14} & 64.00 & \textbf{87.00} & \textbf{80.00} \\
\midrule
\multicolumn{10}{c}{DeepSeek-R1-Distill-Llama-8B} \\ 
\midrule
FullKV & 79.00 & 60.74 & 76.32 & 68.00 & 72.45 & 78.47 & 56.00 & 68.00 & 80.80 \\
H2o & \textbf{74.00} & 60.00 & 67.11 & 72.00 & 72.45 & 70.83 & 52.00 & 61.00 & 65.40 \\
StreamingLLM & 73.00 & 59.26 & 70.39 & 69.00 & 72.45 & 77.08 & 50.00 & \textbf{66.00} & 34.40 \\
PyramidKV & 50.00 & 58.52 & 69.74 & \textbf{73.00} & \textbf{73.21} & 74.31 & 41.00 & 58.00 & 43.00 \\
\rowcolor{gray!10}
\textbf{Lethe(ours)} & 71.00 & \textbf{60.74} & \textbf{75.00} & 72.00 & \textbf{73.21} & \textbf{79.17} & \textbf{54.00} & 63.00 & \textbf{77.80} \\
\midrule
\multicolumn{10}{c}{DeepSeek-R1-Distill-Llama-70B} \\
\midrule
FullKV & 93.00 & 85.93 & 92.11 & 83.00 & 88.30 & 95.83 & 73.00 & 93.00 & 88.80 \\
H2o & 92.00 & 84.44 & 92.76 & \textbf{87.00} & 89.81 & 95.14 & \textbf{70.00} & 82.00 & 80.60 \\
StreamingLLM & \textbf{94.00} & 85.19 & 92.76 & 82.00 & 89.05 & \textbf{95.83} & 67.00 & 89.00 & 75.20 \\
PyramidKV & 83.00 & 84.44 & 91.45 & 82.00 & 89.43 & 94.44 & 62.00 & 78.00 & 77.80 \\
\rowcolor{gray!10}
\textbf{Lethe(ours)} & 87.00 & \textbf{87.41} & \textbf{93.42} & 85.00 & \textbf{90.19} & 95.14 & \textbf{70.00} & \textbf{90.00} & \textbf{82.00} \\
\bottomrule
\end{tabular}
\caption{Performance comparison of Lethe with PyramidKV, H2O, StreamingLLM, and FullKV on math500 and MMLU. Abbreviations: \textit{abs.\ algebra} (abstract algebra), \textit{anat.} (anatomy), \textit{astron.} (astronomy), \textit{bus.\ ethics} (business ethics), \textit{clin.\ knowl.} (clinical knowledge), \textit{coll.\ biol.} (college biology), \textit{coll.\ chem.} (college chemistry), \textit{coll.\ CS} (college computer science).}
\label{tab:performance_comparison}
\end{table*}
}

{
\renewcommand{\arraystretch}{0.55}
\begin{table}[ht]
\centering
\begin{tabular}{lccccc}
\toprule
\multirow{2}{*}{\textbf{Method}} & \multicolumn{5}{c}{\textbf{batch size}} \\
\cmidrule(lr){2-6}
 & 1 & 4 & 8 & 16 & 32 \\
\midrule
\multicolumn{6}{c}{DeepSeek-R1-Distill-Qwen-7B} \\
\midrule
FullKV & 1552 & 10308 & 66504 & 66370 & OOM \\
Lethe(ours) & 1162 & 6372 & 40680 & 66602 & 66624 \\
\midrule
\multicolumn{6}{c}{DeepSeek-R1-Distill-Qwen-32B} \\
\midrule
FullKV & 1751 & 17373 & 18633 & OOM & OOM \\
Lethe(ours) & 1551 & 18617 & 18639 & 18659 & OOM \\
\midrule
\multicolumn{6}{c}{DeepSeek-R1-Distill-Llama-8B} \\
\midrule
FullKV & 1555 & 21062 & 65096 & 65811 & OOM \\
Lethe(ours) & 879 & 2107 & 5383 & 18643 & 65589 \\
\midrule
\multicolumn{6}{c}{DeepSeek-R1-Distill-Llama-70B} \\
\midrule
FullKV & 2096 & 29640 & 36610 & 36694 & OOM \\
Lethe(ours) & 896 & 5110 & 18124 & 36472 & 36602 \\
\bottomrule
\end{tabular}
\caption{Per-GPU generation memory (MB) across models and batch sizes. 70B uses 3$\times$ GPUs due to tensor parallelism. Lethe reduces memory vs. FullKV and avoids OOM.}
\label{tab:memory}
\end{table}
}

\subsection{An Illustration of Lethe’s Layerwise and Temporal KV Cache Pruning}
\label{Illustration}
Figure~\ref{fig_lethe_illustration} illustrates Lethe’s layer- and time-aware pruning. At step 1000, Lethe prunes differently across layers (e.g., layer 13 vs. 27). By step 2000, retained tokens include both recent outputs and previously retained entries. Gray tokens denote evictions; yellow and blue indicate retained tokens from current and prior steps. Lethe reduces attention length from thousands to hundreds of tokens without quality loss, enabling efficient long-context decoding.

\section{Experiment}
\label{Experiment}
\subsection{Experimental Setup}
\label{E-Setup}
We evaluate Lethe from two perspectives: \textbf{accuracy preservation} and \textbf{inference efficiency}. These aspects are examined across a range of model sizes and practical scenarios to assess the generality and robustness of the proposed method.

Our experiments are conducted on four reasoning-oriented language models from the DeepSeek-R1-Distill family: Qwen-7B, Qwen-32B, LLaMA-8B, and LLaMA-70B. All models are tested on NVIDIA A100 80GB GPUs, with 3-way model parallelism applied for the 70B variant. Due to hardware limitations, the full 671B model is omitted; its officially released distilled variants are used instead as proxies.

We adopt two challenging benchmarks: \textbf{Math500}~\cite{lightman2023lets}, a set of complex math problems, and a subset of \textbf{MMLU}~\cite{hendrycks2021ethics} covering 8 diverse subjects. Together, they test both symbolic reasoning and factual understanding. Lethe is compared against four baselines: (i) \textit{FullKV}, which retains the full history of tokens without pruning; (ii) \textit{H2O}~\cite{zhang2023h2o}, a top-$k$ attention-based retention method; (iii) \textit{StreamingLLM}~\cite{xiao2023efficient}, which uses a fixed-size sliding window for cache management; and (iv) \textit{PyramidKV}~\cite{cai2024pyramidkv}, which applies a static layerwise allocation strategy. All baselines are re-implemented within a unified framework to ensure consistency. \textit{SnapKV}~\cite{li2024snapkv} is excluded, as its design centers around long-form input prefill, making it unsuitable for incremental decoding with intermediate reasoning.

Our evaluation covers both single-batch and multi-batch decoding scenarios. In the former, we vary token lengths from 1.5k to 20k to simulate long-form generation. In the latter, we scale the batch size from 1 to 32 to reflect concurrent serving workloads. Key performance metrics include decoding latency, peak memory usage, and token throughput.

Ablation studies are conducted to analyze the influence of Lethe's two main hyperparameters: \texttt{sparse\_ratio}, which determines the threshold for sparsity-based token pruning, and \texttt{recent\_ratio}, which specifies the fraction of most recent tokens preserved irrespective of sparsity. These experiments help to characterize the trade-offs between memory reduction and generation quality.

{
\renewcommand{\arraystretch}{0.55}
\begin{table}[ht]
\centering
\begin{tabular}{lccccc}
\toprule
\multirow{2}{*}{\textbf{Method}} & \multicolumn{5}{c}{\textbf{batch size}} \\
\cmidrule(lr){2-6}
 & 1 & 4 & 8 & 16 & 32 \\
\midrule
\multicolumn{6}{c}{DeepSeek-R1-Distill-Qwen-7B} \\
\midrule
FullKV & 33.1 & 104.1 & 117.7 & 111.0 & OOM \\
Lethe(ours) & 34.6 & 111.7 & 177.3 & 193.3 & 245.0 \\
\midrule
\multicolumn{6}{c}{DeepSeek-R1-Distill-Qwen-32B} \\
\midrule
FullKV & 15.2 & 39.1 & 52.9 & OOM & OOM \\
Lethe(ours) & 15.6 & 55.3 & 85.4 & 91.8 & OOM \\
\midrule
\multicolumn{6}{c}{DeepSeek-R1-Distill-Llama-8B} \\
\midrule
FullKV & 30.1 & 90.9 & 115.6 & 123.4 & OOM \\
Lethe(ours) & 31.4 & 124.2 & 217.8 & 316.4 & 385.7 \\
\midrule
\multicolumn{6}{c}{DeepSeek-R1-Distill-Llama-70B} \\
\midrule
FullKV & 8.3 & 24.1 & 28.7 & 29.4 & OOM \\
Lethe(ours) & 8.5 & 28.7 & 47.1 & 68.7 & 89.5 \\
\bottomrule
\end{tabular}
\caption{Throughput (tokens/s) across models and batch sizes. Lethe consistently outperforms FullKV, especially at larger batches. "OOM" denotes out-of-memory.}
\label{tab:throughput}
\end{table}
}

\begin{figure*}[ht]
\centering
\includegraphics[width=7in]{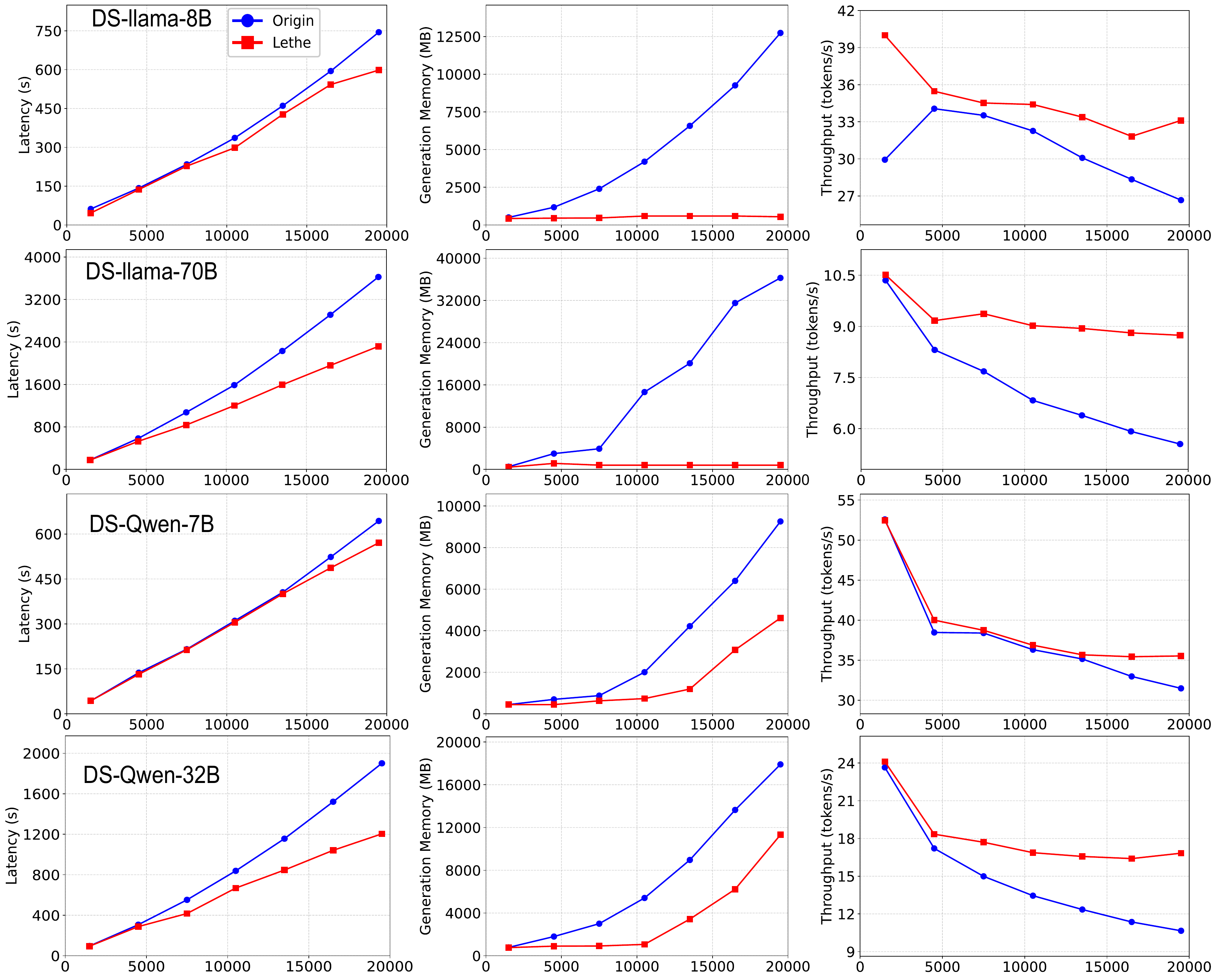}
\caption{Latency, generation memory and throughput performance versus generated tokens comparing full KV cache (blue) and Lethe (red)}
\label{fig_token_level}
\end{figure*}
\subsection{Accuracy Preservation under Cache Compression}
\label{E-Accuracy}

We evaluate Lethe's ability to retain model accuracy under aggressive KV cache pruning. Results are summarized in Table~\ref{tab:performance_comparison}.

Lethe consistently preserves or closely matches the accuracy of FullKV across models and tasks. On the challenging Math500 benchmark, Lethe maintains strong performance: under Qwen-7B, it achieves 85.4\% vs. 86.4\% (FullKV), while H2O and StreamingLLM degrade to 67.2\% and 65.8\%, respectively. Similar trends hold for Qwen-32B and LLaMA-8B, where Lethe outperforms other baselines by significant margins (e.g., $+$17.3\% over StreamingLLM on LLaMA-8B).

On MMLU, Lethe exhibits stable accuracy across diverse subjects. Notably, under Qwen-32B, Lethe slightly surpasses FullKV on categories such as \textit{abstract algebra} and \textit{business ethics}, suggesting that moderate cache pruning may remove noisy context and benefit generalization. StreamingLLM performs poorly in subjects requiring long-range context, such as \textit{philosophy} and \textit{clinical knowledge}, while Lethe maintains strong and balanced accuracy.

Compared to PyramidKV, Lethe avoids the drop in performance caused by static per-layer pruning. For instance, under LLaMA-70B on Math500, PyramidKV suffers a 7.9\% drop relative to FullKV, while Lethe preserves nearly full accuracy (88.1\% vs. 88.7\%). Overall, Lethe strikes a robust balance between compression and accuracy.

\subsection{Inference Efficiency: Latency, Memory and Throughput Gains}
\label{E-Inference Efficiency}

To assess the inference performance of our proposed method, \textit{Lethe}, we systematically compare it with the original baseline across four representative large language models: DeepSeek-R1-Distill-Qwen-7B, DeepSeek-R1-Distill-Qwen-32B, DeepSeek-R1-Distill-LLaMA-8B, and DeepSeek-R1-Distill-LLaMA-70B. All experiments were conducted on NVIDIA A100 GPUs, with the 70B model evaluated using three A100 GPUs in parallel. The generation memory is computed as the peak GPU memory usage minus the memory immediately after model loading.

\subsubsection{Batch-Level Performance}

Lethe significantly reduces memory usage across all batch sizes and prevents out-of-memory (OOM) errors that occur with FullKV at batch size 32. Throughput is consistently improved; for example, on LLaMA-70B at batch size 32, Lethe enables inference (89.5 tok/s), while FullKV fails.

\subsubsection{Token-Level Scalability}

Figures~\ref{fig_token_level} show Lethe’s latency, memory, and throughput trends over longer token lengths. Lethe reduces latency by 20–40\% and achieves a stable memory footprint even at 20k tokens. For LLaMA-70B, memory usage plateaus at ~800MB post-6k tokens, compared to 36GB+ for FullKV.

\subsection{Ablation Study}

We fix \texttt{sparse\_ratio}=400 and \texttt{recent\_ratio}=0.3 unless otherwise noted. Increasing \texttt{sparse\_ratio} improves accuracy but with diminishing returns beyond 400. Lower values harm performance due to over-pruning.

Similarly, \texttt{recent\_ratio}=0.3 yields a strong balance between coherence and compression. Higher values retain unnecessary tokens, while smaller ones risk breaking contextual flow.

\section{Conclusion}

In this paper, we proposed \textbf{Lethe}, an adaptive KV cache compression framework to improve the inference efficiency of large language models. Unlike prior work that targets static prompts, Lethe focuses on the overlooked overhead from intermediate tokens in multi-step reasoning tasks. Empirical analysis reveals both \textit{spatial heterogeneity} and \textit{temporal dynamics} in attention patterns during decoding. To address this, Lethe employs layer-wise sparsity estimation for adaptive per-layer cache allocation and an LRU-inspired mechanism for dynamic token pruning. Experiments on Math500 and MMLU show that Lethe maintains accuracy while achieving up to \textbf{2.56$\times$} inference speedup, highlighting its potential for efficient and scalable LLM deployment.

\section{Acknowledgments}
This work was supported in part by the Shaanxi Key Technology R\&D Program under Grant 2024GX-ZDCYL-02-15 and the Natural Science Funds for Distinguished Young Scholar of Shaanxi under Grant 2025JC-JCQN-079.

\bibliography{aaai2026}

\end{document}